\documentclass[reprint,amsmath,amssymb,aps,floatfix]{revtex4-2}
\usepackage{graphicx}
\usepackage{dcolumn}
\usepackage{bm}
\usepackage{hyperref}
\usepackage{lipsum}
\usepackage{subcaption}  
\usepackage{caption}         
\usepackage{float}
\usepackage{tikz}
\usepackage{ragged2e}
\usepackage{multirow}
\usepackage{placeins}
\usetikzlibrary{positioning, arrows.meta}  
\usepackage{relsize}  
\usepackage{overpic}  
\usepackage{placeins}
\usepackage{amssymb}
\usepackage{amsmath}
\usepackage{algpseudocode}
\usepackage{makecell}
\usepackage{bm}
\usepackage{placeins}
\usepackage[export]{adjustbox}
\usepackage{amsthm}

\theoremstyle{remark}

\usepackage{mathtools}
\usepackage{float}
\usepackage[linesnumbered,ruled,vlined]{algorithm2e}
\usetikzlibrary{positioning, arrows.meta}  
\usepackage{relsize}  
\usepackage{overpic}

\usepackage{tabularx}
\usepackage{booktabs}
\usepackage{array}
\usepackage[ruled,vlined]{algorithm2e}

\begin{document}

\preprint{APS/123-QED}

\title{Learning from Noise: Effective-Rank Collapse and Out-of-Distribution Rejection \\ in Restricted Boltzmann Machines}

\author{Oshada Rathnayake}
\email{fcu8ss@virginia.edu}
\author{Nikhil Shukla}
\affiliation{%
University of Virginia, Charlottesville, VA, 22903, USA
}%

\begin{abstract}

Restricted Boltzmann machines (RBMs) represent data by shaping an energy landscape over visible and hidden configurations, but their discriminative use is fragile under out-of-distribution (OOD) inputs: samples outside the training distribution can be absorbed into one of the learned class basins rather than rejected. Here, we analyze this failure mode through the spectrum of the induced visible--visible interaction \(J=WW^{T}\), where \(W\) is the visible--hidden weight matrix. Relative to a Marchenko--Pastur random-matrix reference, conventional training spreads spectral weight into many weak, bulk-compatible directions, increasing the effective rank of \(J\). When auxiliary random binary images are assigned to a rejection label during training, the learned interaction undergoes effective-rank collapse: weak bulk-like modes are depleted, spectral weight concentrates into fewer dominant eigendirections, and the effective rank of \(J\) approaches that of the empirical data covariance matrix. The resulting RBM rejects structured OOD image datasets while preserving MNIST classification accuracy, showing that random auxiliary exposure can reshape both the interaction spectrum and the free-energy landscape of an energy-based classifier.

\end{abstract}

\maketitle

\section{Introduction}
Restricted Boltzmann machines (RBMs) \cite{smolensky1986information} are energy-based probabilistic models~\cite{lecun2006tutorial} in which learning corresponds to shaping an energy landscape over visible and hidden configurations. After training, configurations consistent with the data distribution are assigned low free-energy, whereas statistically incompatible configurations are expected to lie at higher free-energy~\cite{hinton2012practical,krizhevsky2009learning}. This representation allows RBMs to capture latent statistical structure in data and has supported their use in both generative modeling and discriminative inference~\cite{le2011learning,fischer2014training,schmah2008generative}. 

\begin{figure}
    \centering
    \includegraphics[width=1\linewidth]{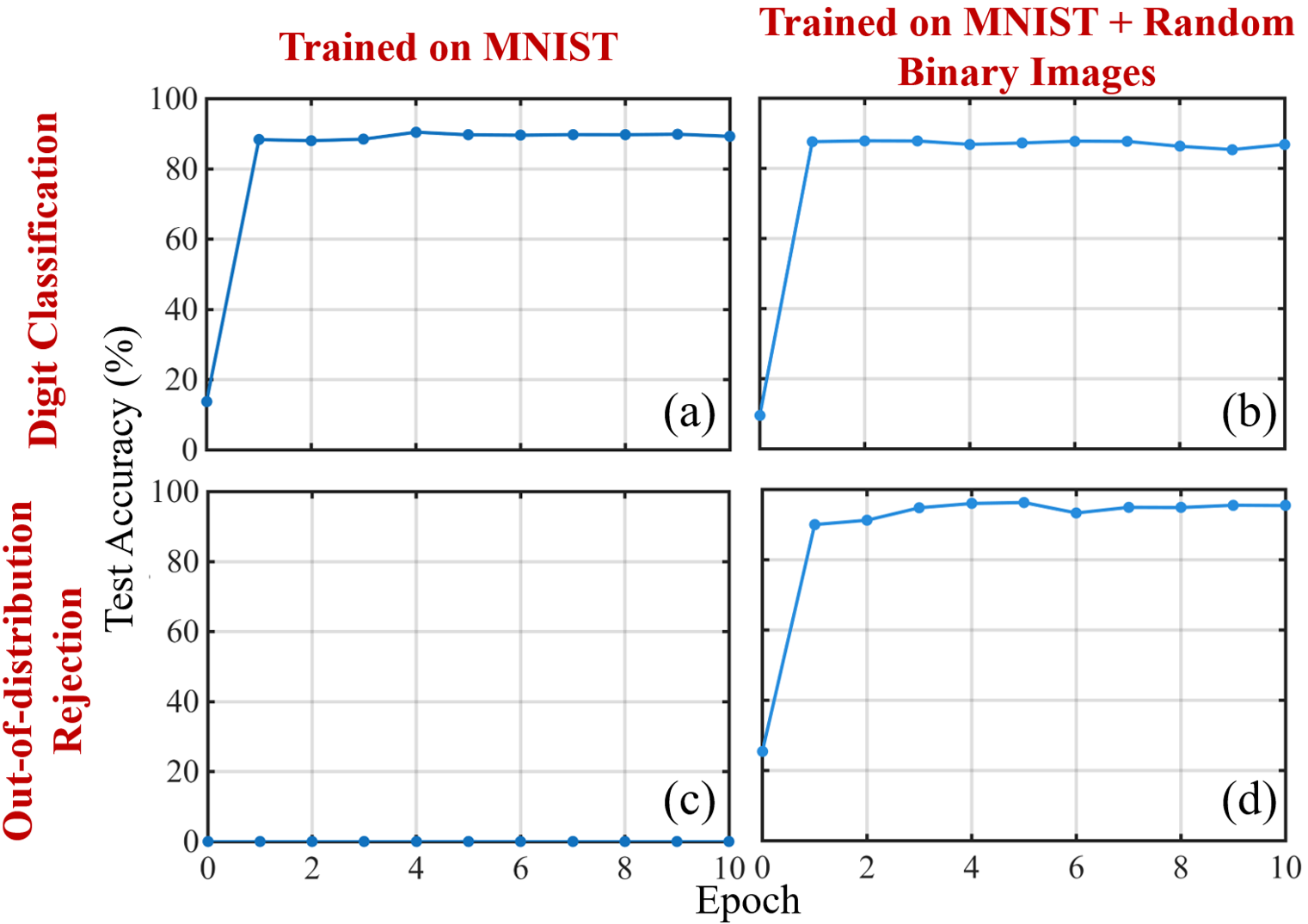}
    \caption{\justifying  Comparison of the digit classification and out-of-distribution (OOD) rejection performance of an RBM trained exclusively on MNIST~\cite{lecun1998gradient} and an RBM trained on MNIST + random binary images. Panels (a) and (b) show the digit classification accuracy on the MNIST test set, demonstrating comparable in-distribution performance for both models. Panels (c) and (d) show the OOD rejection performance on the KMNIST dataset \cite{clanuwat2018deep}. Unlike the conventional RBM, the auxiliary-trained RBM successfully assigns the majority of OOD samples to the rejection class while preserving its digit classification performance.}
    \label{fig:test accuracy MNIST and KMNIST}
\end{figure}

In this work, we focus on the latter setting. As shown in Fig.~\ref{fig:test accuracy MNIST and KMNIST}(a), a conventionally trained RBM can achieve high classification accuracy on in-distribution data; similar results have been shown in prior work~\cite{larochelle2008classification}  as well. However, the same model exhibits a persistent limitation when evaluated on out-of-distribution (OOD) inputs~\cite{santos2024adversarial, hendrycks2017a,lee2018simple}. Since the classifier selects the lowest-energy label among the learned alternatives, an input far from the training distribution can still be absorbed into one of the in-distribution class basins rather than ``rejected". In the baseline model considered here, this behavior leads to vanishing OOD rejection accuracy, as shown in Fig.~\ref{fig:test accuracy MNIST and KMNIST}(c). This failure mode
exposes a structural limitation of the learned energy landscape: accurate classification within the training distribution does not, by itself, guarantee the formation of a separate rejection region for inputs outside that distribution.

A direct way to introduce a rejection state is to augment the training set with auxiliary OOD    samples~\cite{hendrycks2018deep}. In this work, these auxiliary samples are chosen to be random binary images, with each pixel independently drawn from a Bernoulli distribution with probability $1/2$, and are assigned to an additional rejection label. As shown in Fig.~\ref{fig:test accuracy MNIST and KMNIST}(d), this training protocol enables the RBM to assign OOD inputs to the rejection class, while preserving the in-distribution digit-classification performance shown in Fig.~\ref{fig:test accuracy MNIST and KMNIST}(b). 
\

The central objective of this work to provide insight to the aforementioned behavior of RBMs through the effective interactions induced by the hidden layer. Since the RBM has no direct
visible--visible couplings, correlations among visible units are mediated
through the visible--hidden weights. The leading pairwise signature of this
mediation is the induced visible--visible interaction
\begin{equation}
\label{eq:J}
J = W W^{T},
\end{equation}
where \(W\) is the visible--hidden weight matrix. Thus, although the RBM is bipartite, \(J\) provides a compact second-order projection of the learned visible-space structure. Its spectrum gives a basis-independent diagnostic of the number and relative strength of the directions used by the hidden layer to represent the data.  

Under conventional training, the effective rank of \(J\) increases steadily, indicating that spectral weight is distributed over an increasing number of weakly occupied directions. A substantial fraction of these directions remains close to the Marchenko--Pastur bulk~\cite{marvcenko1967distribution} associated with an unstructured random weight matrix of comparable scale, and is therefore difficult to distinguish from noise-dominated structure. These weak spectral directions provide additional low-significance channels through which OOD inputs can overlap with the learned in-distribution manifold and be assigned to one of the digit classes.

Training with auxiliary random binary images suppresses this expansion into bulk-like directions. The resulting spectrum undergoes what we refer to as \emph{rank collapse}: spectral weight is depleted from weak, noise-dominated directions and concentrated into a smaller number of dominant eigendirections that separate from the random-matrix bulk. This reduction in rank is not a uniform loss of learned structure. The dominant modes that separate from the random-matrix bulk are retained, while spectral weight in weak bulk-like directions is suppressed. As a result, the effective rank of \(J\) is driven toward the effective rank of the empirical data covariance matrix. The observed rank collapse therefore reflects a reduction to the dimensional scale supported by the data, rather than an over-regularized low-rank limit. In contrast, conventional training continues to populate low-significance directions beyond this covariance reference, producing additional bulk-like channels that are not required for in-distribution classification.

The spectral reorganization is accompanied by a corresponding change in the learned energy landscape. In the conventionally trained RBM, different OOD families occupy distinct energy levels and can be absorbed into separate class-dependent basins. When auxiliary random binary images are included during training, these OOD energy levels collapse toward a common low-energy region associated with the rejection label. The same change is reflected dynamically in the faster decay of the energy autocorrelation during Gibbs sampling, which is consistent with a less rugged effective landscape. Thus, auxiliary exposure modifies the RBM at both the spectral and energetic levels: weak bulk-like directions in \(J=WW^{T}\) are suppressed, while diverse OOD inputs are collected into a coherent rejection region. This combined reorganization enables OOD rejection without degrading in-distribution classification accuracy.

We now describe the RBM architecture and training protocols used to compare conventional training with training that includes auxiliary random-binary-image exposure.

\section{Network Architecture and Training}
\label{sec:Network}

We use the MNIST dataset~\cite{lecun1998gradient} as the in-distribution dataset. Each MNIST image is represented as a binarized \(28\times 28\) pixel array, giving \(N_{v_x}=784\) pixel-visible units. To perform classification within the RBM, the class label is appended to the visible layer as a one-hot vector. The visible layer therefore contains
\(
N_v = N_{v_x}+N_{v_y}=784+11=795
\)
units, where the first ten label units encode the MNIST digit classes \(0,\ldots,9\), and the eleventh label unit denotes the OOD rejection class. The hidden layer contains \(N_h=100\) binary hidden units. The models are trained
using one-step contrastive divergence (CD-1)~\cite{carreira2005contrastive}, in which the negative phase is
approximated by a single Gibbs update. Finally, a batch size of 1 is used for the training experiments~\cite{hinton2012practical}. This architecture and the hyperparameters are kept fixed throughout all experiments, so that differences in performance and spectral structure can be attributed to the training protocol rather than to changes in model capacity. Sensitivity analyses with respect to
the hidden-layer size \(N_h\) and the number of contrastive-divergence steps $k$ are
reported in Appendix~\ref{app:architecture_cd}.

We compare two training protocols. In the baseline protocol, the RBM is trained exclusively on the \(60{,}000\) MNIST training samples, with the appropriate digit label activated for each image and the OOD label inactive. In the auxiliary-exposure protocol, each training epoch also includes random binary images. For these auxiliary samples, each pixel is drawn independently from a Bernoulli distribution with probability \(1/2\), and the OOD label unit is activated while all digit-label units are set to zero. Unless otherwise stated, \(5{,}000\) random binary images are added per epoch; we also examine other auxiliary-sample sizes to determine how the amount of exposure affects rejection performance. The auxiliary samples therefore define a controlled, unstructured OOD class against which the MNIST-only baseline can be compared.

\section{Spectrum of the Learned Visible--Visible Interaction}
\label{sec:spectral_structure_rbm}

We first characterize the learned representation through the spectrum of the induced visible--visible interaction \(J=WW^{T}\).

\subsection{Effective Rank of the Induced Interaction}
\label{subsec:effective_rank}

We begin by quantifying the spectral dimensionality of the interaction induced by
the hidden layer. Let
\begin{equation}
    W \in \mathbb{R}^{N_v \times N_h}
\end{equation}
denote the learned visible--hidden weight matrix, with \(W_{i\mu}\) the coupling
between visible unit \(i\) and hidden unit \(\mu\). As introduced above, the
corresponding induced visible--visible interaction is
\(
    J = W W^{T},
    \label{eq:J_WWT}
\)
or, equivalently,
\(
    J_{ij} = \sum_{\mu=1}^{N_h} W_{i\mu} W_{j\mu}.
    \label{eq:J_WWT_components}
\)
Thus, \(J_{ij}\) is large when visible units \(i\) and \(j\) couple to the hidden
layer through similar weight profiles.

The matrix \(J\) is symmetric positive semi-definite, and its rank satisfies
\begin{equation}
    \mathrm{rank}(J)
    =
    \mathrm{rank}(W W^{T})
    \leq
    \min(N_v,N_h).
    \label{eq:J_rank_bound}
\end{equation}
Equivalently, the non-zero eigenvalues of \(J\) are the squared singular values
of \(W\). The algebraic rank, however, is sensitive to small singular values and numerical
fluctuations~\cite{falini2022review}, and therefore may not provide a robust measure of the effective
dimensionality of the learned representation. We therefore use the entropy-based effective rank~\cite{roy2007effective}, which assigns weights to spectral-modes in accordance to their contribution to the total spectral power.

Let

\begin{equation}
    0 \leq \lambda_{1} \leq \lambda_{2} \leq \cdots \leq \lambda_{N_v}
\end{equation}
denote the eigenvalues of \(J\). We define the normalized spectral weights
\begin{equation}
    p_i
    =
    \frac{\lambda_i}{\sum_{j=1}^{N_v} \lambda_j},
    \qquad
    \sum_{i=1}^{N_v} p_i = 1,
    \label{eq:spectral_weights}
\end{equation}
and the associated spectral entropy~\cite{shannon1948mathematical}
\begin{equation}
    S_{\mathrm{spec}}
    =
    -\sum_{i=1}^{N_v} p_i \log p_i .
    \label{eq:spectral_entropy}
\end{equation}
The effective rank is then
\begin{equation}
    r_{\mathrm{eff}}
    =
    \exp\!\left(S_{\mathrm{spec}}\right).
    \label{eq:effective_rank}
\end{equation}
This quantity provides a smooth measure of the number of spectral directions that carry appreciable weight. If the spectrum is distributed over many comparable modes, \(r_{\mathrm{eff}}\) is large; if the spectral weight is concentrated in a small number of dominant modes, \(r_{\mathrm{eff}}\) is small. 

\begin{figure}[h]
    \centering
    \includegraphics[width=1\linewidth]{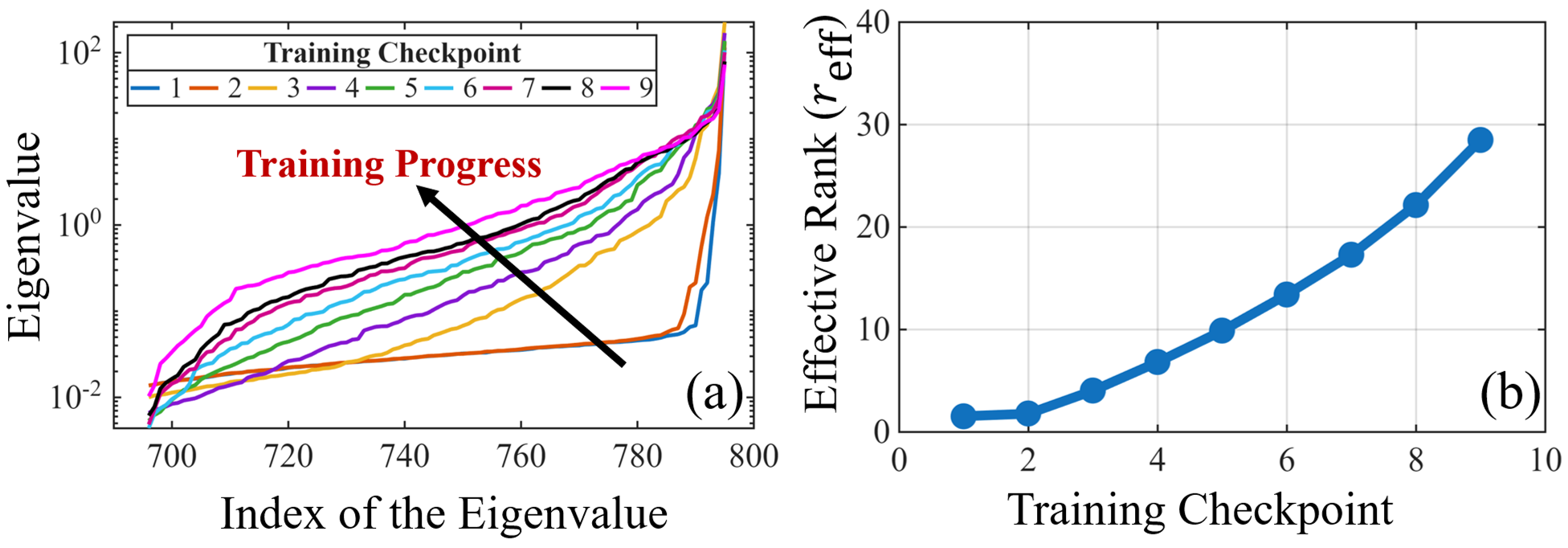}
    \caption{\justifying (a) Eigenvalue spectra of \( J = W W^T \), with the indices of the eigenvalues, sorted in descending order, at training checkpoints, which are selected when the model reaches training accuracies between 10\% and 90\% in 10\% increments, with training accuracy monitored after each parameter update (60,000 updates per epoch). The evolution of the spectrum illustrates the progressive distribution of spectral weight during learning. (b) Corresponding evolution of the effective rank (\(r_{\mathrm{eff}}\)) of \( J = W W^T \) evaluated at the same training checkpoints, showing the growth of the effective spectral dimensionality throughout training.}
    \label{fig:MNIST_100_eig_idx}
\end{figure}

Fig.~\ref{fig:MNIST_100_eig_idx}(a) shows the evolution of the eigenvalue spectrum of \(J\) during conventional MNIST training. As training progresses, spectral weight is transferred into modes that are initially weakly occupied, broadening the nonzero spectrum. This redistribution increases \(r_{\mathrm{eff}}\) as shown in Fig.~\ref{fig:MNIST_100_eig_idx}(b), indicating an expansion of the effective dimensionality of the visible-space representation.

\

\subsection{Marchenko--Pastur Null Model and Signal--Noise Separation}
\label{subsec:mp}

The entropy-based effective rank quantifies the spread of spectral weight, but is unable to distinguish the resolved structure from directions that are compatible with random fluctuations. To facilitate this distinction, we compare the spectrum of the induced interaction \(J=WW^{T}\) with the random-matrix spectrum expected for an unstructured weight matrix.

Let \(\lambda_i\) denote the eigenvalues of \(J\), and define the scaled eigenvalues
\begin{equation}
  \tilde{\lambda}_i = \frac{\lambda_i}{N_h}.
  \label{eq:scaled_eigs}
\end{equation}
If the entries of \(W\) are independent random variables with zero mean and variance \(\sigma^2\), then the empirical spectral density of \(N_h^{-1}J\) converges, in the limit \(N_v,N_h\to\infty\) at fixed aspect ratio \(Q=N_v/N_h\), to the Marchenko--Pastur distribution~\cite{marvcenko1967distribution}. For \(Q>1\), this distribution consists of a continuous bulk supported on \([\lambda_-,\lambda_+]\), with
\begin{equation}
  \lambda_{\pm}
  =
  \sigma^2\left(1 \pm \sqrt{Q}\right)^2 ,
  \label{eq:mp_edges}
\end{equation}
together with a point mass of weight \(1-1/Q\) at zero, reflecting the
rank deficiency of \(J\). In the architecture considered here,
\(N_v=795\) and \(N_h=100\), so \(Q\simeq 7.95\). Thus, approximately
\(1-N_h/N_v\simeq 0.87\) of the eigenvalues are structurally zero, while the
non-zero random bulk of the scaled spectrum lies in
\([3.31\,\sigma^2,14.6\,\sigma^2]\).

\begin{figure*}[t]
    \centering
    \includegraphics[width=1\linewidth]{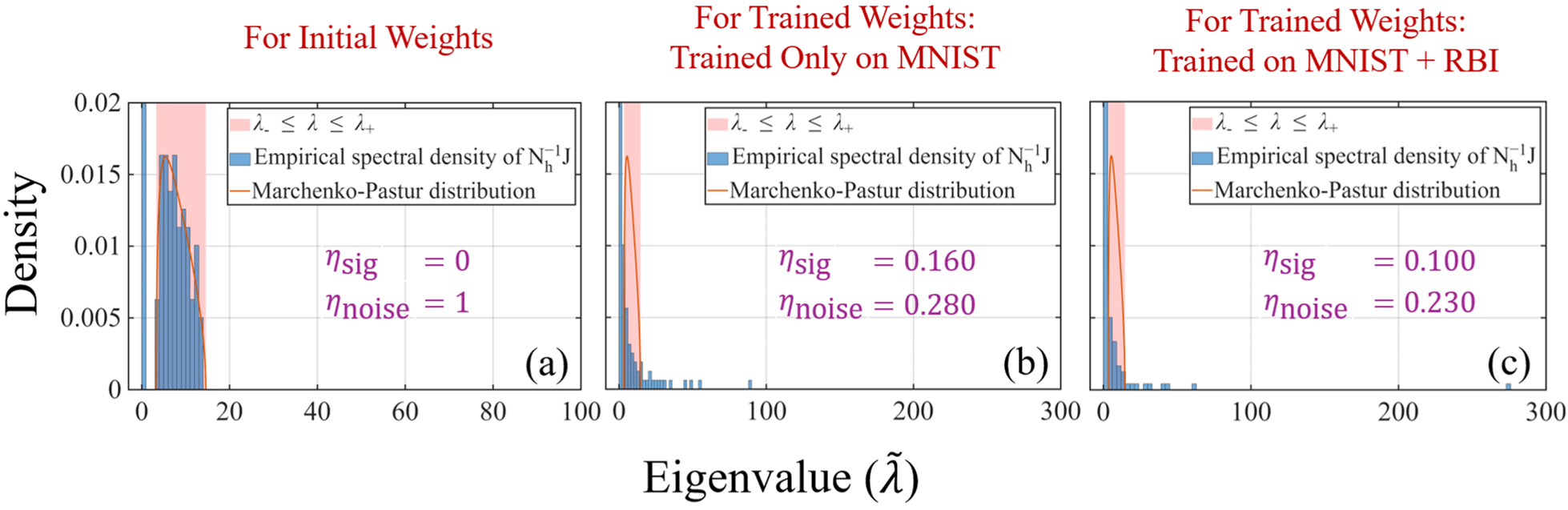}
    \caption{\justifying Comparison between the Marchenko–Pastur distribution and the empirical spectral density of \(N_h^{-1}J\) corresponding to (a) the randomly initialized initial weight matrix, (b) the learned weight matrix of the RBM trained exclusively on MNIST data, and (c) the learned weight matrix of the RBM trained on both MNIST and random binary images. Inclusion of auxiliary data yields fewer eigenvalues beyond and within the Marchenko–Pastur bulk, indicating a reduction in informative-feature-dominated directions as well as noise-dominated directions leading to the \textit{rank collapse}. For clarity, the vertical axis is truncated, and the full heights of the truncated bars are 0.874 in (a), 0.465 in (b), and 0.318 in (c).}
    \label{fig:random_matrix_analysis}
\end{figure*}

The upper edge \(\lambda_+\) provides a reference scale for separating
bulk-compatible directions from spectral outliers. Eigenvalues satisfying
\(\lambda_- \leq \tilde{\lambda}_i \leq \lambda_+\) are consistent with the
unstructured null model, whereas eigenvalues with
\(\tilde{\lambda}_i>\lambda_+\) have separated from the random bulk~\cite{plerou2002random}. We therefore
define the fraction of available modes that appear as signal-like outliers,
\begin{equation}
  \eta_{\mathrm{sig}}
  =
  \frac{
  \#\left\{ i:\tilde{\lambda}_i>\lambda_+ \right\}
  }{\min(N_v,N_h)} ,
  \label{eq:eta_sig}
\end{equation}
and the fraction of available modes that remain within the random-matrix bulk,
\begin{equation}
  \eta_{\mathrm{noise}}
  =
  \frac{
  \#\left\{ i: \lambda_- \leq \tilde{\lambda}_i \leq \lambda_+ \right\}
  }{\min(N_v,N_h)} .
  \label{eq:eta_noise}
\end{equation}
Here \(\eta_{\mathrm{noise}}\) should be interpreted as the fraction of
bulk-compatible, noise-dominated directions rather than as a direct measure of
label noise or data corruption.

At initialization, the entries of the visible--hidden weight matrix are drawn
independently from a zero-mean Gaussian distribution,
\(
    W_{i\mu}\sim \mathcal{N}(0,10^{-4})
\). Before learning, \(J=WW^{T}\) is therefore a Wishart-type random matrix. As shown in Fig.~\ref{fig:random_matrix_analysis}(a), the empirical spectrum of \(N_h^{-1}J\) is well described by the Marchenko--Pastur distribution, up to finite-size fluctuations. This agreement is consistent with the absence of resolved spectral structure beyond that expected from an unstructured random weight matrix.

After training, the entries of \(W\) are no longer independent, and learning
also changes the overall weight scale. We therefore treat the parameter
\(\sigma^2\) in Eq.~\eqref{eq:mp_edges} as an effective bulk variance for the
trained spectrum, rather than as the initialization variance. With this
interpretation, the Marchenko--Pastur distribution serves as a random-matrix
reference for bulk-compatible spectral weight; it is not assumed that the
trained spectrum is exactly Marchenko--Pastur distributed.

Figs.~\ref{fig:random_matrix_analysis}(b) and
\ref{fig:random_matrix_analysis}(c) compare the empirical spectrum of
\(N_h^{-1}J\) after training with this reference for the two training
protocols. In both cases, a subset of eigenvalues separates above the upper
Marchenko--Pastur edge \(\lambda_+\), identifying directions that are resolved
relative to the random null model. A second subset remains within the interval
\([\lambda_-,\lambda_+]\), corresponding to bulk-compatible directions, while
eigenvalues below \(\lambda_-\) carry negligible spectral weight and are
effectively inactive. Compared with the MNIST-only RBM, the RBM trained with
auxiliary random binary images contains fewer bulk-compatible modes and a larger
fraction of inactive directions. Thus, auxiliary exposure suppresses weak
spectral directions rather than merely rescaling the spectrum. This redistribution
provides the random-matrix signature of the effective-rank collapse analyzed in
Sec.~\ref{subsec:rank_collapse}.

\subsection{Rank Collapse and Spectral Concentration}
\label{subsec:rank_collapse}

\begin{figure*}[t]
    \centering
    \includegraphics[width=.9\linewidth]{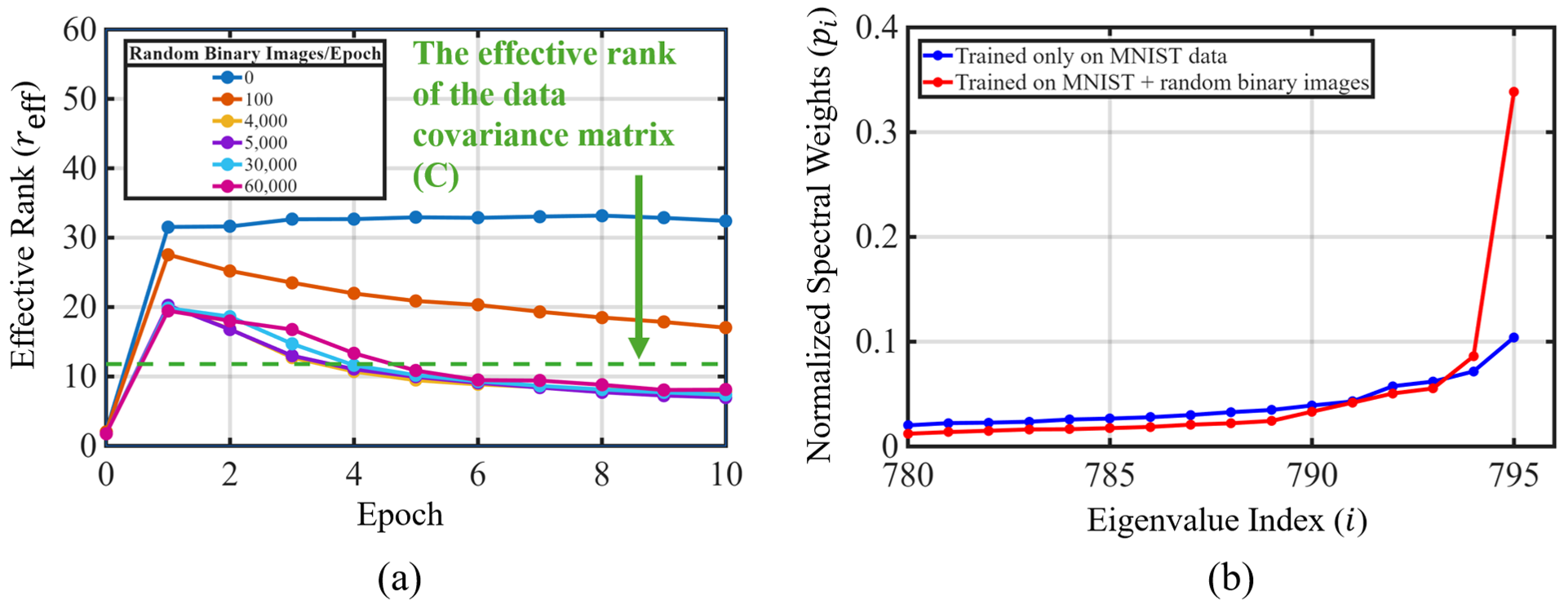}
    \caption{\justifying (a) Evolution of the effective rank of \( J = W W^T \) during training for different numbers of random binary images incorporated per epoch. The case labeled (0) corresponds to the RBM trained exclusively on MNIST data, whereas all other cases correspond to RBMs trained on both MNIST data and random binary images. (b) Comparison of normalized spectral weights of eigendirections of \(J\) obtained after training with the two different methods.}    
    \label{fig:MNIST_Rank_concentration}
\end{figure*}

We next compare the evolution of \(r_{\mathrm{eff}}\) under the two training protocols. At each epoch, \(r_{\mathrm{eff}}\) is computed from the spectrum of the induced interaction \(J=WW^{T}\). As shown in
Fig.~\ref{fig:MNIST_Rank_concentration}(a), the RBM trained only on MNIST shows a monotonic increase in effective rank before reaching a plateau. Conventional training therefore distributes spectral weight over an increasing number of visible-space directions. By contrast, when auxiliary random binary images are included during training, \(r_{\mathrm{eff}}\) decreases after the initial stage of learning and saturates at a substantially lower value. We refer to this reduction of the spectral dimensionality of \(J\) as effective-rank collapse.
The Marchenko--Pastur comparison in Sec.~\ref{subsec:mp} shows that this collapse is not a uniform contraction of the spectrum. Rather, auxiliary exposure preferentially suppresses weak directions that remain compatible with
the random-matrix bulk, while retaining a smaller set of dominant modes that carry most of the spectral weight. Thus, the reduced effective rank should not be interpreted simply as a loss of learned features. It reflects a redistribution of spectral weight away from weak bulk-compatible directions and toward a more concentrated set of leading eigendirections.
\begin{figure}[b]
    \centering
    \includegraphics[width=0.9\linewidth]{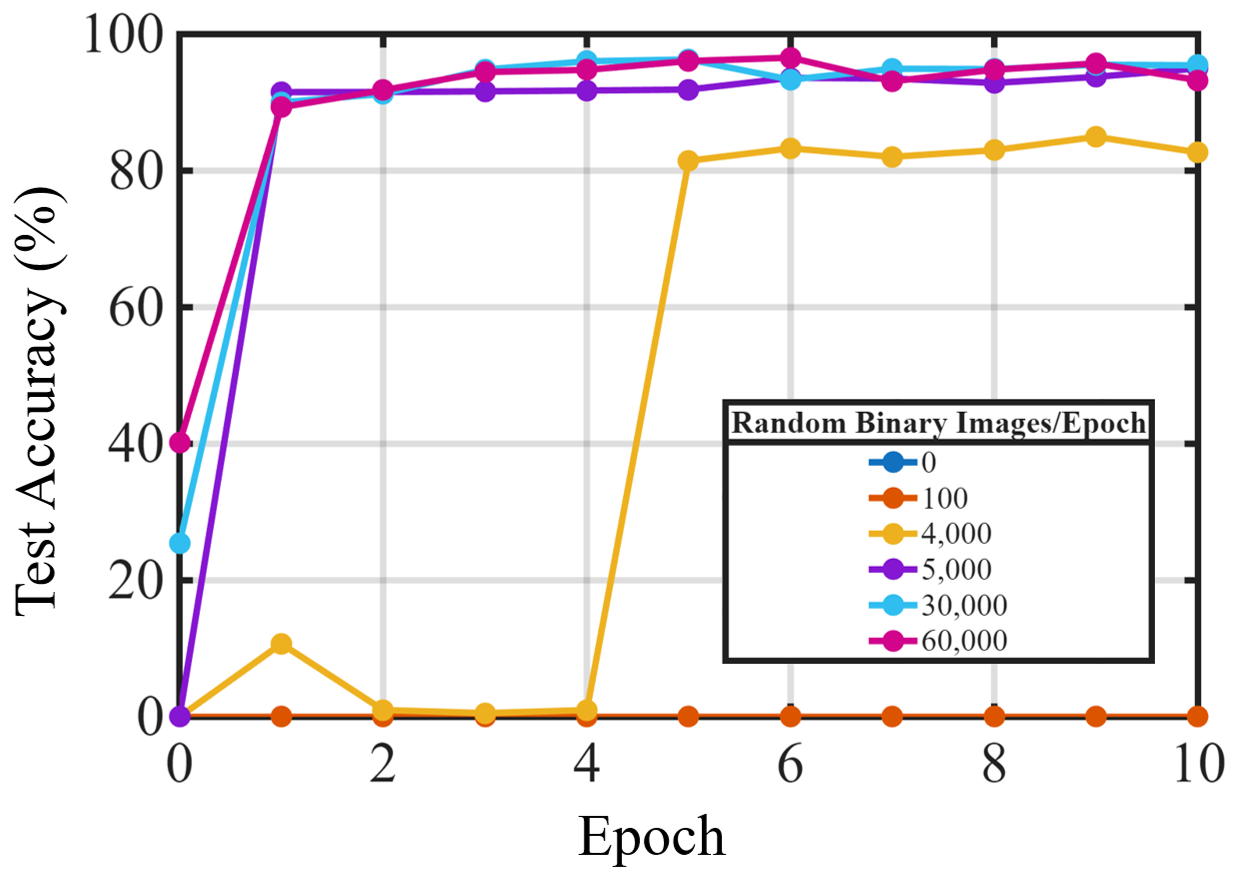}
    \caption{\justifying Out-of-distribution (OOD) rejection performance on the KMNIST dataset~\cite{clanuwat2018deep} for RBMs trained with different numbers of auxiliary random binary images per training epoch. The legend entry 0 corresponds to the conventional RBM trained exclusively on MNIST (i.e., without auxiliary random binary images). Incorporating sufficient auxiliary random binary images enables the RBM to identify OOD samples and assign them to the rejection class, and the OOD rejection accuracy increases with the number of auxiliary random binary images used during training before reaching a plateau, indicating saturation of the acquired OOD rejection capability.
}
    \label{fig:OOD_rejection}
\end{figure} 
This concentration is shown directly in Fig.~\ref{fig:MNIST_Rank_concentration}(b), which compares the normalized
spectral weights \(p_i\) defined in Eq.~\eqref{eq:spectral_weights}. The auxiliary-trained RBM places a larger fraction of the total spectral weight in the leading modes and exhibits a strongly suppressed weak tail. In contrast, the MNIST-only RBM assigns appreciable weight to a broader set of low-weight modes, many of which lie close to the bulk-compatible spectral region identified in Sec.~\ref{subsec:mp}. The lower value of \(r_{\mathrm{eff}}\) in the auxiliary-trained model therefore reflects spectral concentration rather than an
overall depletion of the learned representation.

A useful reference scale is provided by the empirical data covariance matrix \(C\)~\cite{krizhevsky2009learning}, discussed in Appendix~\ref{app:analogy_J_dataCovariance}. As shown in Fig.~\ref{fig:MNIST_Rank_concentration}(a), auxiliary exposure drives \(r_{\mathrm{eff}}(J)\) toward \(r_{\mathrm{eff}}(C)\). We use \(r_{\mathrm{eff}}(C)\) as a data-dependent estimate of the dimensional scale supported by the training distribution. From this perspective, the MNIST-only RBM expands beyond this scale by populating additional low-significance directions, whereas the auxiliary-trained RBM remains closer to the spectral dimensionality of the data itself.

The functional behavior associated with this spectral reorganization is shown in Figs.~\ref{fig:test accuracy MNIST and KMNIST} and~\ref{fig:OOD_rejection}. The decrease in \(r_{\mathrm{eff}}\) does not degrade the MNIST classification accuracy, which remains approximately \(89\%\). At the same time, the OOD rejection accuracy increases with the number of auxiliary random binary images included during training and eventually reaches a plateau. These results are consistent with the following interpretation: weak bulk-compatible directions provide additional channels through which OOD inputs can overlap with the learned digit representation, while auxiliary exposure suppresses these directions and concentrates the representation into the dominant modes needed for in-distribution classification. Effective-rank collapse therefore marks a selective reorganization of the learned interaction matrix, rather than a capacity-limited degradation of the RBM.

\section{Free-Energy Landscape and Gibbs-Sampling Dynamics}
\label{sec:energy}
The spectral analysis of Sec.~\ref{sec:spectral_structure_rbm} characterizes the
learned RBM through the induced interaction \(J=WW^{T}\). We now examine the
same two training protocols from the perspective of the free-energy landscape.
This provides a visible-space description of where in-distribution and
out-of-distribution inputs are placed after the hidden variables have been
marginalized.

\begin{figure}[b]
    \centering
    \includegraphics[width=1\linewidth]{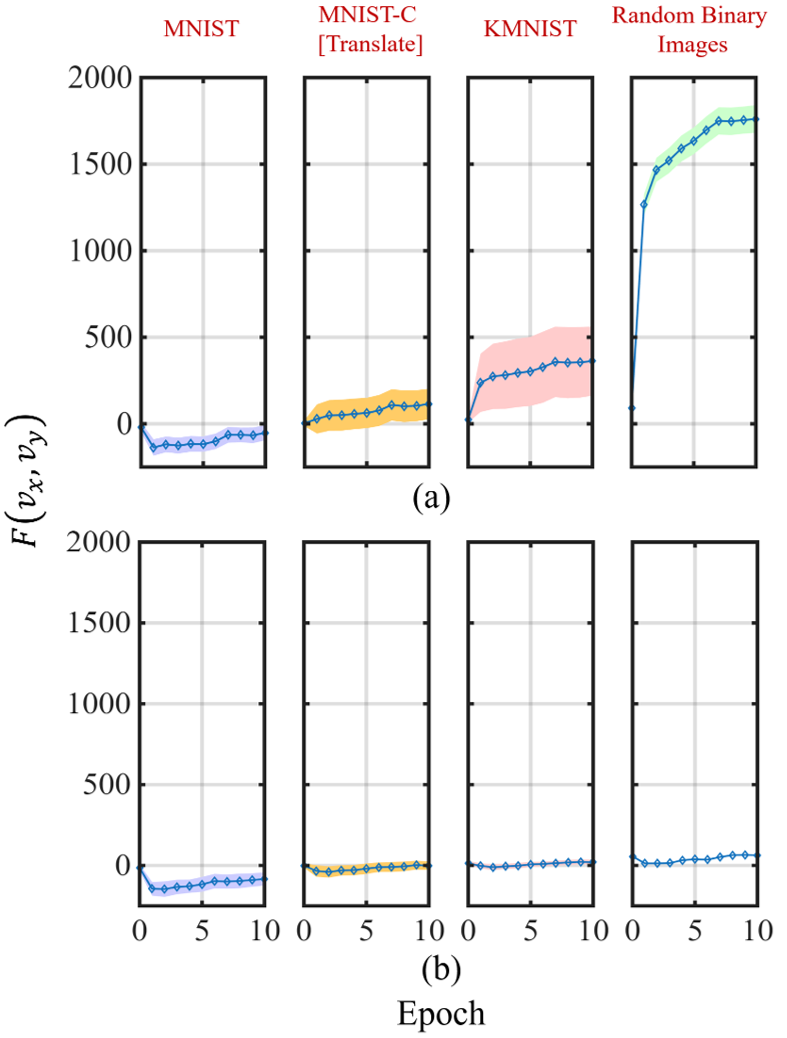}
    \caption{\justifying Mean free-energy and free-energy variance of MNIST samples, samples from two distinct OOD datasets, and random binary image samples evaluated at different training epochs (a) for the RBM trained exclusively on MNIST data and (b) for the RBM trained with MNIST data + random binary images.}
    \label{fig:free_enegy}
\end{figure}

The visible layer is partitioned into pixel units
\(v_x \in \{0,1\}^{N_{v_x}}\) and label units
\(v_y \in \{0,1\}^{N_{v_y}}\). These units are coupled to hidden units
\(h \in \{0,1\}^{N_h}\) through weight blocks
\(W_x \in \mathbb{R}^{N_{v_x}\times N_h}\) and
\(W_y \in \mathbb{R}^{N_{v_y}\times N_h}\). The full visible--hidden weight
matrix is
\begin{equation}
    W = [\,W_x;\,W_y\,] \in \mathbb{R}^{N_v\times N_h},
    \qquad
    N_v=N_{v_x}+N_{v_y}.
\end{equation}
The RBM is defined by the joint energy
\begin{equation}
  E(v_x,v_y,h)
  =
  -a^{T}v_x
  -b^{T}v_y
  -c^{T}h
  -v_x^{T}W_x h
  -v_y^{T}W_y h ,
  \label{eq:joint_energy}
\end{equation}
where \(a\), \(b\), and \(c\) are the pixel, label, and hidden biases,
respectively~\cite{hinton2012practical}. Since the hidden variables are latent, dataset-level comparisons
are made using the free energy obtained by summing over \(h\),
\begin{equation}
  F(v_x,v_y)
  =
  -\log \sum_h \exp[-E(v_x,v_y,h)] .
  \label{eq:free_energy_definition}
\end{equation}
Since the hidden units are conditionally independent, this marginalization can
be carried out analytically, giving

\begin{equation}
\begin{split}
  F(v_x,v_y)
  =&
  -a^{T}v_x
  -b^{T}v_y\\
  &-
  \sum_{\mu=1}^{N_h}
  \log\!\left[
  1+
  \exp\!\left(
  c_\mu
  +
  (W_x^{T}v_x)_\mu
  +
  (W_y^{T}v_y)_\mu
  \right)
  \right].
  \label{eq:free_energy}
  \end{split}
\end{equation}

Thus, each labeled visible configuration \((v_x,v_y)\) is assigned a unique
free energy, independent of any particular sampled hidden state~\cite{hinton2012practical}.
\begin{figure}[b]
    \centering
    \includegraphics[width=1\linewidth]{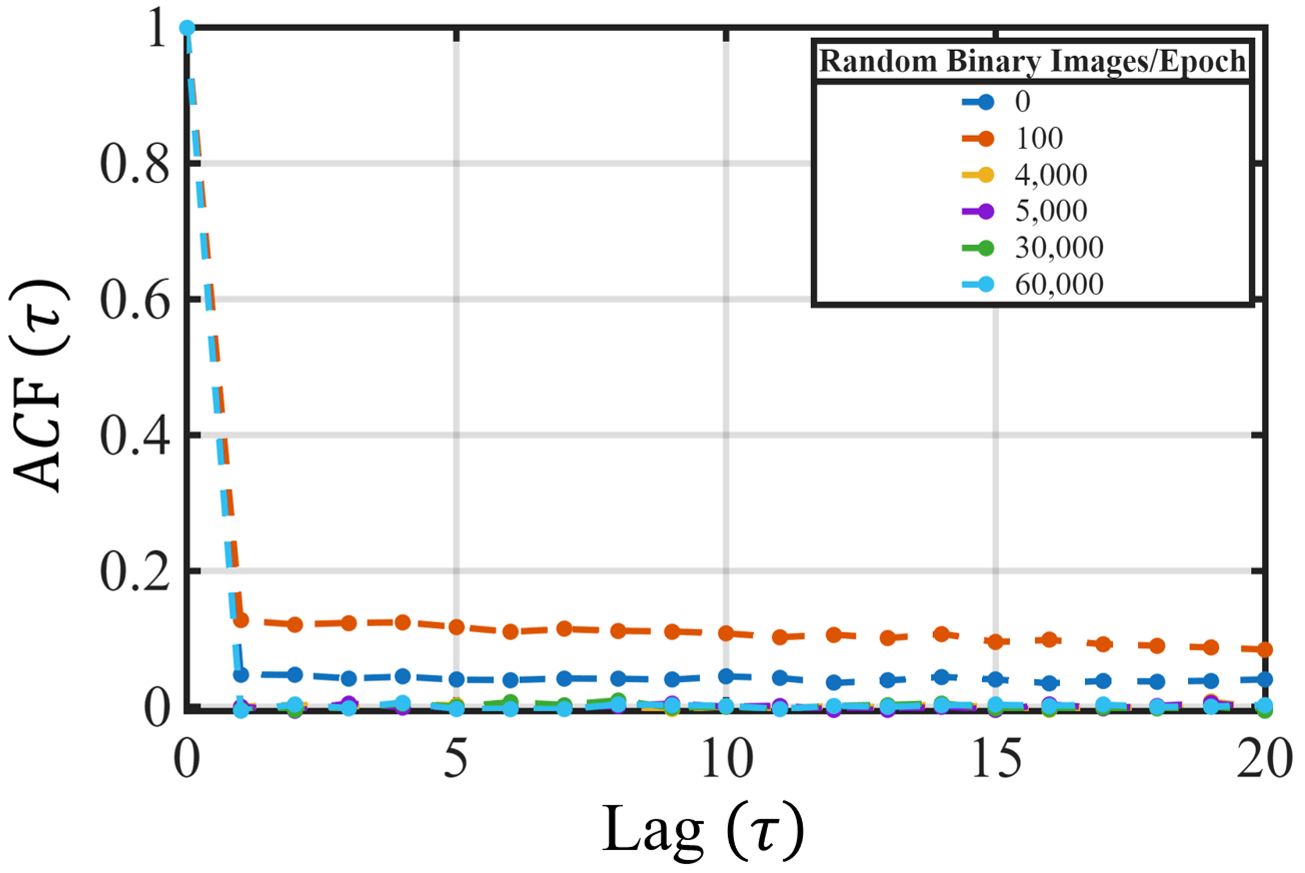}
    \caption{\justifying Autocorrelation decay of free-energy trajectories obtained from Gibbs sampling
for RBMs trained with different numbers of auxiliary random binary images per
epoch. The curve labeled \(0\) corresponds to the RBM trained exclusively on
MNIST, whereas the remaining curves correspond to RBMs trained on MNIST together
with auxiliary random binary images. The MNIST-only model and the model trained
with \(100\) random binary images per epoch exhibit slower autocorrelation
decay, indicating longer memory of the sampled free-energy level. By contrast,
models trained with \(4000\) or more random binary images per epoch show faster
decorrelation, consistent with the reduced fragmentation of the free-energy
landscape associated with the emergence of OOD rejection.}
    \label{fig:ACF}
\end{figure}
\begin{figure*}
    \centering
    \includegraphics[width=1\linewidth]{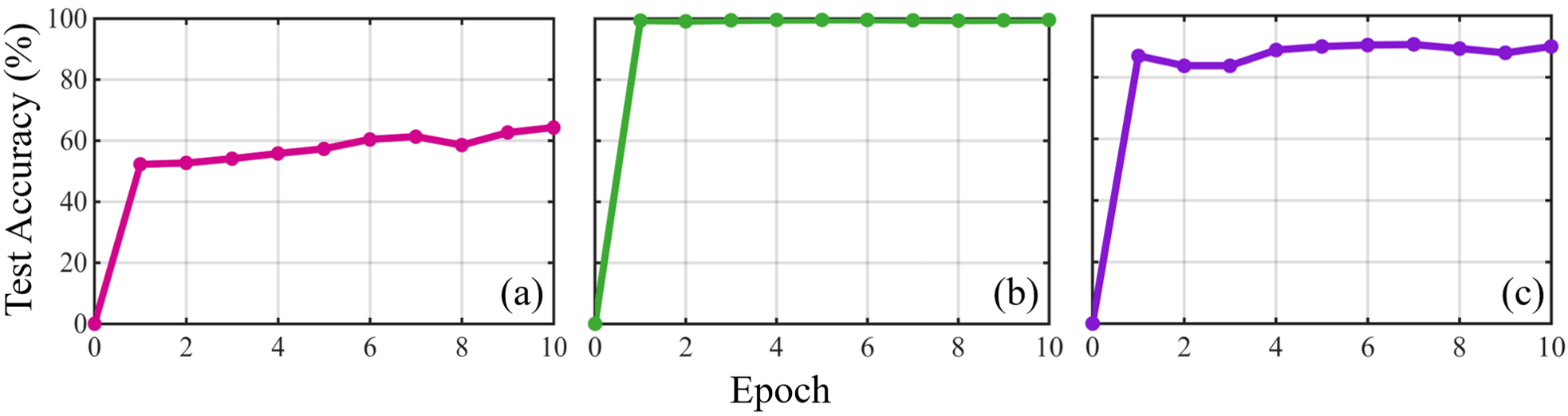}
    \caption{\justifying Out-of-distribution (OOD) rejection performance of the RBM trained on MNIST~\cite{lecun1998gradient} with auxiliary random binary images. Accuracy of identifying (a) MNIST-C (translate)~\cite{mu2019mnist}, (b) notMNIST~\cite{flanagan_notmnist_mnist}, and (c) Fashion-MNIST~\cite{xiao2017fashion}(c) as OOD data.}
    \label{fig:example}
\end{figure*}
We first use \(F(v_x,v_y)\) as a static diagnostic of the learned landscape. For
MNIST samples, \(v_y\) is set to the corresponding digit label. For OOD samples, \(v_y\) is
set to the OOD rejection label. For each evaluation dataset \(\mathcal{D}\), we
compute the distribution of
\(\{F(v_x^{(n)},v_y^{(n)})\}_{n=1}^{N_{\mathcal{D}}}\) during training and
summarize it by its mean and variance at each epoch.

For the RBM trained only on MNIST, Fig.~\ref{fig:free_enegy}(a) shows that MNIST samples occupy a comparatively low-free-energy region throughout training. By contrast, OOD samples evaluated with the rejection label remain at higher
free energies. Thus, in the MNIST-only model, the rejection label is not associated with a low-free-energy region for these OOD inputs.

The behavior changes when auxiliary random binary images are included during training. As shown in Fig.~\ref{fig:free_enegy}(b), the free-energy level of MNIST samples remains comparable to that of the MNIST-only model, whereas the OOD datasets and random binary images are shifted toward lower free energies under the rejection label. Auxiliary exposure therefore reorganizes the free-energy landscape so that heterogeneous OOD inputs are more favorably assigned to the rejection state, rather than to the in-distribution digit labels.

We next probe the landscape dynamically using block Gibbs sampling. Starting from the trained RBM, we alternate between \(h \sim p(h\,|\,v)\) and \(v \sim p(v\,|\,h)\), where \(v=(v_x,v_y)\) \cite{roussel2021barriers}. Along the resulting trajectory, we record the free energy of the visible state,
\begin{equation}
  X_t = F(v_{x,t},v_{y,t}).
\end{equation}
The normalized autocorrelation is then defined as
\begin{equation}
  ACF(\tau)
  =
  \frac{
  \langle X_t X_{t+\tau}\rangle-\langle X\rangle^2
  }{
  \langle X^2\rangle-\langle X\rangle^2
  } .
  \label{eq:autocorr}
\end{equation}
A slowly decaying autocorrelation indicates that the Gibbs trajectory retains
memory of its free-energy level over many updates, as expected when the
landscape contains separated or metastable regions. Faster decay is consistent
with a less fragmented effective landscape~\cite{box2015time}. Fig.~\ref{fig:ACF} compares the free-energy autocorrelation \(ACF(\tau)\) obtained from Gibbs-sampling trajectories for the different training protocols. The MNIST-only RBM, as well as RBMs trained with very few auxiliary random binary images, exhibits a slower decay of \(ACF(\tau)\), indicating that the sampler retains memory of its free-energy level over longer times. By contrast, RBMs trained with sufficiently many auxiliary random binary images decorrelate more rapidly. This faster decay is consistent with a less fragmented effective free-energy landscape after auxiliary exposure.

Together, the static free-energy diagnostic and the Gibbs-sampling autocorrelation indicate that auxiliary random-binary-image exposure reorganizes the RBM landscape: OOD inputs become energetically more favorable under the rejection label, while the low-free-energy structure associated with MNIST samples is preserved.

\section{Out-of-Distribution Rejection Across Image Datasets}
\label{sec:example}

Beyond the KMNIST~\cite{clanuwat2018deep} comparison shown in Fig.~\ref{fig:test accuracy MNIST and KMNIST}, we use the same auxiliary-trained RBM to evaluate OOD rejection on additional image datasets. MNIST remains the in-distribution dataset, and the only auxiliary OOD samples used during training are random binary images assigned to the rejection label. The trained model is evaluated on MNIST-C (translate) \cite{mu2019mnist}, notMNIST~\cite{flanagan_notmnist_mnist}, and Fashion-MNIST \cite{xiao2017fashion}, which provide structured non-MNIST inputs with different visual statistics. The MNIST-C (translate) dataset is obtained by applying spatial translation distortions to MNIST images. Such translations pose a significant challenge for RBMs, causing a substantial degradation (20\% of accuracy) in digit classification performance on the MNIST-C (translate) dataset. Therefore, MNIST-C (translate) data is treated as out-of-distribution samples in this study, notMNIST consists of glyphs of the letters A--J drawn from different fonts, and Fashion-MNIST consists of images of clothing items.

For each OOD test sample, successful rejection corresponds to activation of the eleventh label unit. Fig.~\ref{fig:example} shows the rejection accuracy as a function of training epoch for MNIST-C (translate), notMNIST, and Fashion-MNIST. In all three cases, the RBM trained with auxiliary random binary images successfully assigns OOD samples to the rejection class, although the rejection performance varies across datasets. These results demonstrate that the acquired OOD rejection capability generalizes to diverse OOD distributions while maintaining high classification accuracy for the in-distribution data.

\section{Conclusion}
\label{sec:conclusion}

In this work, we have shown that auxiliary random-binary-image exposure changes the learned
structure of an RBM classifier in a way that enables out-of-distribution
rejection. In the MNIST-only baseline, unfamiliar inputs are absorbed into one
of the learned digit-class basins. When random binary images are assigned to the
rejection label during training, the RBM instead develops a populated rejection
state while preserving MNIST classification accuracy.

This change is reflected in the spectrum of the induced interaction
\(J=WW^{T}\). Conventional training spreads spectral weight into weak,
bulk-compatible directions, increasing the effective rank of \(J\). Auxiliary
exposure suppresses this weak spectral tail and concentrates the representation
into fewer dominant eigendirections, driving the effective rank toward that of
the empirical data covariance matrix. The same training reorganizes the
free-energy landscape so that heterogeneous OOD inputs are collected into a
common rejection region. The faster decay of free-energy autocorrelations under
Gibbs sampling is consistent with this reduced landscape fragmentation. Thus, OOD rejection in this setting is associated with
a joint spectral and free-energy reorganization of the RBM.\\\\

\begin{flushleft}
\textbf{DATA AVAILABILITY:} The data that support the findings of this study are available from the corresponding author upon request.\
\end{flushleft}

\begin{flushleft}
\textbf{CODE AVAILABILITY:} The codes associated with this manuscript are available from the corresponding author upon request.
\end{flushleft}
\appendix

\section{Sensitivity to Hidden Layer Size \((N_h)\) and Contrastive Divergence \((CD-k)\)}
\label{app:architecture_cd}

\subsection*{Sensitivity to the Number of Hidden Units}
\label{subsec:hidden_sensitivity}

We evaluate the dependence on hidden-layer size by training RBMs with \(N_h=20,50,100,200,\) and \(400\) under the two protocols used in the main text: MNIST-only training and MNIST training with auxiliary random binary images.

\begin{figure*}
    \centering
    \includegraphics[width=0.9\linewidth]{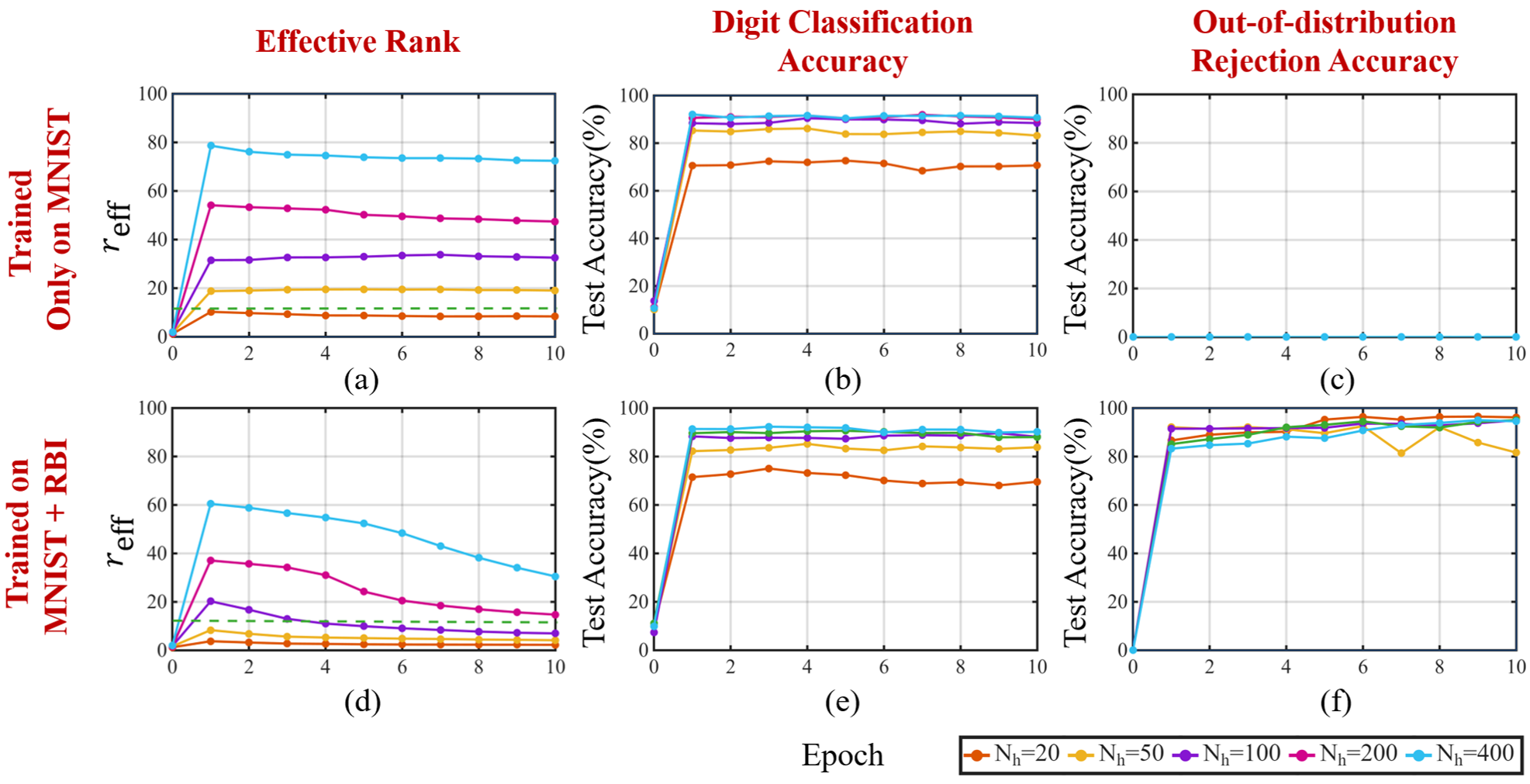}
    \caption{\justifying Sensitivity to the number of hidden units \(N_h\). Panels (a)--(c) correspond to RBMs trained exclusively on MNIST, while panels (d)--(f) correspond to RBMs trained on MNIST with auxiliary random binary images. Panels (a) and (d) show the effective rank of the induced interaction \(J=WW^{T}\); the green dashed line marks the effective rank of the empirical data covariance matrix \(C\). Panels (b) and (e) show the MNIST digit-classification accuracy, and panels (c) and (f) show the OOD rejection accuracy. The legend applies to all panels.} 
    \label{fig:N_h}
\end{figure*}

Figs.~\ref{fig:N_h}(a) and \ref{fig:N_h}(d) show that the effective rank of \(J=WW^{T}\) increases with \(N_h\), as expected from the larger number of available hidden-layer directions. The corresponding MNIST classification accuracies, shown in Figs.~\ref{fig:N_h}(b) and \ref{fig:N_h}(e), also improve with \(N_h\) before approaching a plateau. Increasing the hidden-layer size therefore primarily increases representational capacity until further hidden units provide only marginal gains in digit classification.

Despite the lower values of \(r_{\mathrm{eff}}\) obtained for smaller hidden layers, the RBM trained exclusively on MNIST does not acquire OOD rejection capability. As shown in Fig.~\ref{fig:N_h}(c), the rejection accuracy remains approximately zero for all values of \(N_h\). Thus, a reduced effective rank arising from limited hidden-layer capacity is distinct from the effective-rank collapse induced by auxiliary random binary images in Sec.~\ref{subsec:rank_collapse}. In the former case, the smaller rank reflects a restricted set of available representational directions and is accompanied, for small \(N_h\), by reduced digit-classification performance. In the latter case, auxiliary exposure suppresses weak bulk-compatible eigendirections while retaining the dominant modes needed for digit classification. The effective rank is therefore reduced without eliminating the informative structure required for MNIST recognition, and high OOD rejection accuracy is obtained, as shown in Fig.~\ref{fig:N_h}(f).

For RBMs trained with auxiliary random binary images, the OOD rejection accuracy remains high over a broad range of hidden-layer sizes once the model has sufficient capacity to classify MNIST accurately. This behavior indicates that the emergence of the rejection state is not controlled by \(N_h\) alone, but by the spectral organization produced by the auxiliary training protocol.

\subsection*{Sensitivity to Contrastive-Divergence Steps}
\label{subsec:cd_sensitivity}

\begin{figure*}
    \centering
    \includegraphics[width=0.9\linewidth]{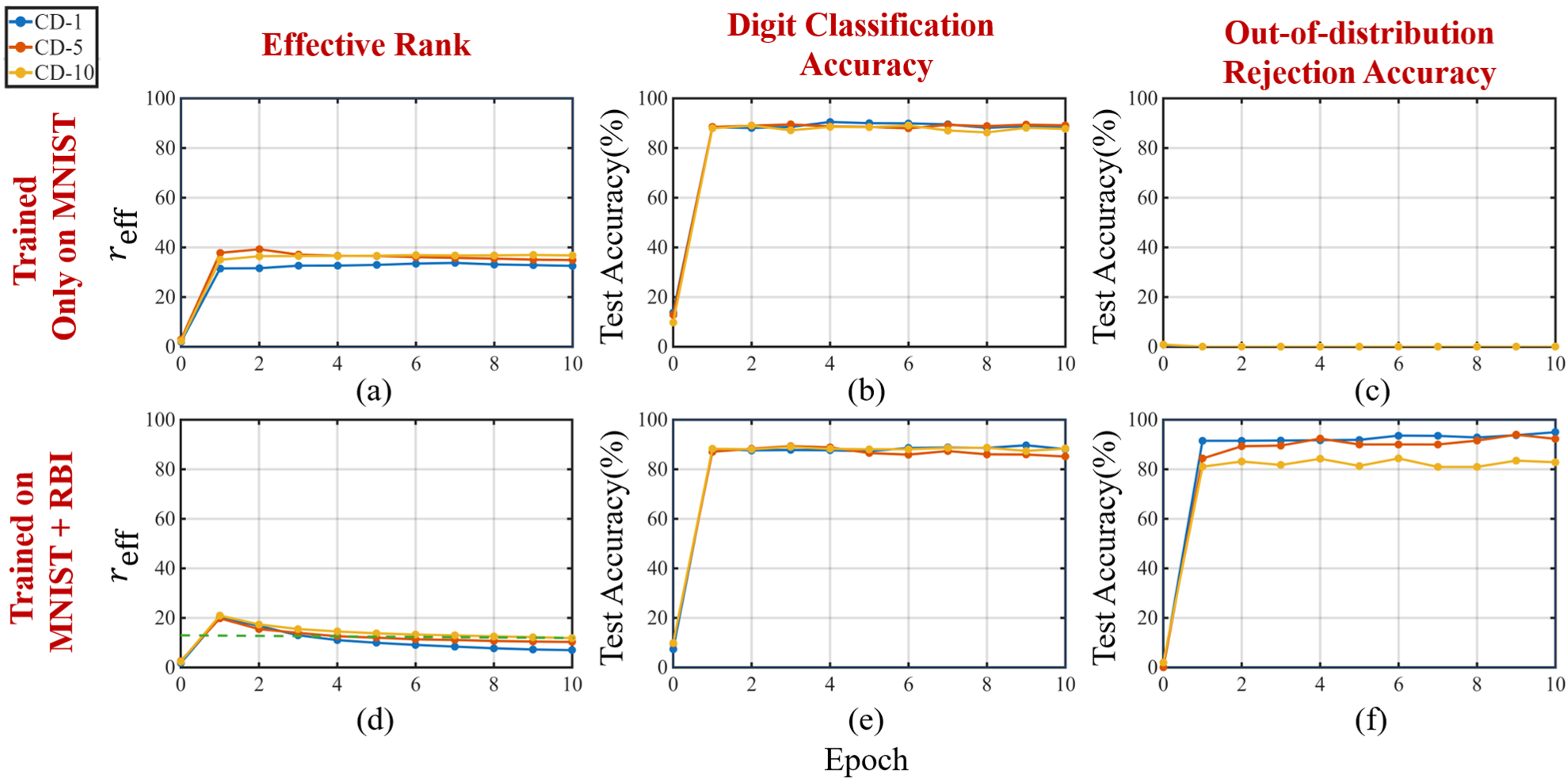}
    \caption{\justifying Sensitivity to the number of Gibbs sampling steps \(k\) used in
contrastive-divergence training. Panels (a)--(c) correspond to RBMs trained exclusively on MNIST, while panels (d)--(f) correspond to RBMs trained on MNIST with auxiliary random binary images. Panels (a) and (d) show the effective rank of the induced interaction \(J=WW^{T}\); the green dashed line marks the effective rank of the empirical data covariance matrix \(C\). Panels (b) and (e) show the MNIST digit-classification accuracy, and panels (c) and (f) show the OOD rejection accuracy. The legend applies to all panels. } 
    \label{fig:CD}
\end{figure*}

We also examine the dependence on the number of Gibbs sampling steps \(k\) used
in contrastive-divergence training. Specifically, we compare CD-\(1\), CD-\(5\),
and CD-\(10\) under the same two training protocols considered above.

Figs.~\ref{fig:CD}(a),(b),(d), and
(e) show that both the effective rank of \(J=WW^{T}\) and the
MNIST digit-classification accuracy depend only weakly on \(k\) over the range
tested. Although larger \(k\) generally provides a better approximation to the
negative phase of the RBM learning rule~\cite{fischer2014training}, this does
not lead to substantial changes in the discriminative performance for the
MNIST classification task considered here. In this setting, even CD-\(1\)
appears sufficient to learn the dominant structures required for digit
classification, and increasing \(k\) produces only marginal changes in
\(r_{\mathrm{eff}}\) and MNIST accuracy.

The OOD rejection behavior shows a similar distinction between the two training
protocols. For the RBM trained exclusively on MNIST, the rejection accuracy
remains approximately zero for all values of \(k\), as shown in
Fig.~\ref{fig:CD}(c). Thus, increasing the number of contrastive-divergence
steps does not by itself produce a rejection state in the MNIST-only model.

For the RBM trained with auxiliary random binary images, Fig.~\ref{fig:CD}(f)
shows that OOD rejection remains high for all tested values of \(k\), although
the rejection accuracy is somewhat lower for larger \(k\). This trend suggests
that the auxiliary-induced rejection state is robust to the contrastive-divergence
parameter, but that the detailed balance between digit-class and rejection
regions can depend on the sampling depth used during training. We therefore use
CD-\(1\) in the main experiments, since it is sufficient to obtain high
in-distribution accuracy, effective-rank collapse, and strong OOD rejection in
the present RBM setting.

\section{Empirical Data Covariance as a Reference Scale}
\label{app:analogy_J_dataCovariance}

In the main text, the effective rank of the empirical data covariance matrix is used as a reference scale for the spectral dimensionality of the learned interaction \(J=WW^{T}\). Here we clarify the basis for this comparison.

Let \(\mathbf{v}\in\{0,1\}^{N_v}\) denote a visible configuration containing both pixel and label units. The empirical covariance matrix of the training data is
\begin{equation}
    C
    =
    \left\langle
    \left(\mathbf{v}-\langle \mathbf{v}\rangle\right)
    \left(\mathbf{v}-\langle \mathbf{v}\rangle\right)^{T}
    \right\rangle_{\mathrm{data}},
\end{equation}
where \(\langle\cdot\rangle_{\mathrm{data}}\) denotes an average over the
training set. The spectrum of \(C\) therefore provides a data-dependent measure
of the number of visible-space directions carrying appreciable second-order
variation in the training distribution.

The RBM contains no direct visible--visible couplings. Nevertheless, the hidden
layer induces an effective second-order interaction among visible units through
the visible--hidden weight matrix \(W\). A compact representation of this
induced interaction is
\begin{equation}
    J = WW^{T}
      = \sum_{\mu=1}^{N_h} \mathbf{w}_{\mu}\mathbf{w}_{\mu}^{T},
\end{equation}
where \(\mathbf{w}_{\mu}\) is the vector of weights connecting hidden unit
\(\mu\) to the visible layer. Thus, each hidden unit contributes a rank-one
visible-space direction, and the spectrum of \(J\) measures how the RBM
allocates representational weight across such directions.

The matrix \(C\) is computed from the training data, whereas \(J=WW^{T}\) is
computed from the learned RBM weights. Since both are \(N_v\times N_v\) matrices
over the visible variables, their spectra provide comparable measures of
visible-space dimensionality. We therefore use \(r_{\mathrm{eff}}(C)\) as a
data-dependent reference scale for the spectral dimensionality supported by the training distribution.

\bibliography{references}

@article{lecun1998gradient,
  title={Gradient-based learning applied to document recognition},
  author={LeCun, Yann and Bottou, L{\'e}on and Bengio, Yoshua and Haffner, Patrick},
  journal={Proceedings of the IEEE},
  volume={86},
  number={11},
  pages={2278--2324},
  year={1998},
  publisher={Ieee}
}

@article{clanuwat2018deep,
  title={Deep learning for classical japanese literature},
  author={Clanuwat, Tarin and Bober-Irizar, Mikel and Kitamoto, Asanobu and Lamb, Alex and Yamamoto, Kazuaki and Ha, David},
  journal={arXiv preprint arXiv:1812.01718},
  year={2018}
}

@article{mu2019mnist,
  title={Mnist-c: A robustness benchmark for computer vision},
  author={Mu, Norman and Gilmer, Justin},
  journal={arXiv preprint arXiv:1906.02337},
  year={2019}
}

@article{xiao2017fashion,
  title={Fashion-mnist: a novel image dataset for benchmarking machine learning algorithms},
  author={Xiao, Han and Rasul, Kashif and Vollgraf, Roland},
  journal={arXiv preprint arXiv:1708.07747},
  year={2017}
}

@inproceedings{roy2007effective,
  title={The effective rank: A measure of effective dimensionality},
  author={Roy, Olivier and Vetterli, Martin},
  booktitle={2007 15th European signal processing conference},
  pages={606--610},
  year={2007},
  organization={IEEE}
}

@article{marvcenko1967distribution,
  title={Distribution of eigenvalues for some sets of random matrices},
  author={Mar{\v{c}}enko, Vladimir A and Pastur, Leonid Andreevich},
  journal={Mathematics of the USSR-Sbornik},
  volume={1},
  number={4},
  pages={457--483},
  year={1967}
}

@book{box2015time,
  title={Time series analysis: forecasting and control},
  author={Box, George EP and Jenkins, Gwilym M and Reinsel, Gregory C and Ljung, Greta M},
  year={2015},
  publisher={John Wiley \& Sons}
}

@techreport{smolensky1986information,
  title={Information processing in dynamical systems: Foundations of harmony theory},
  author={Smolensky, Paul},
  year={1986}
}

@article{le2011learning,
  title={Learning a generative model of images by factoring appearance and shape},
  author={Le Roux, Nicolas and Heess, Nicolas and Shotton, Jamie and Winn, John},
  journal={Neural Computation},
  volume={23},
  number={3},
  pages={593--650},
  year={2011},
  publisher={MIT Press One Rogers Street, Cambridge, MA 02142-1209, USA journals-info~…}
}

@article{schmah2008generative,
  title={Generative versus discriminative training of RBMs for classification of fMRI images},
  author={Schmah, Tanya and Hinton, Geoffrey E and Small, Steven and Strother, Stephen and Zemel, Richard},
  journal={Advances in neural information processing systems},
  volume={21},
  year={2008}
}

@incollection{hinton2012practical,
  title={A practical guide to training restricted Boltzmann machines},
  author={Hinton, Geoffrey E},
  booktitle={Neural Networks: Tricks of the Trade: Second Edition},
  pages={599--619},
  year={2012},
  publisher={Springer}
}

@misc{krizhevsky2009learning,
  title={Learning multiple layers of features from tiny images},
  author={Krizhevsky, Alex and Hinton, Geoffrey and others},
  year={2009},
  publisher={Toronto, ON, Canada}
}

@article{fischer2014training,
  title={Training restricted Boltzmann machines: An introduction},
  author={Fischer, Asja and Igel, Christian},
  journal={Pattern Recognition},
  volume={47},
  number={1},
  pages={25--39},
  year={2014},
  publisher={Elsevier}
}

@article{roussel2021barriers,
  title={Barriers and dynamical paths in alternating Gibbs sampling of restricted Boltzmann machines},
  author={Roussel, Cl{\'e}ment and Cocco, Simona and Monasson, R{\'e}mi},
  journal={Physical Review E},
  volume={104},
  number={3},
  pages={034109},
  year={2021},
  publisher={APS}
}

@inproceedings{carreira2005contrastive,
  title={On contrastive divergence learning},
  author={Carreira-Perpinan, Miguel A and Hinton, Geoffrey},
  booktitle={International workshop on artificial intelligence and statistics},
  pages={33--40},
  year={2005},
  organization={PMLR}
}

@article{hendrycks2018deep,
  title={Deep anomaly detection with outlier exposure},
  author={Hendrycks, Dan and Mazeika, Mantas and Dietterich, Thomas},
  journal={arXiv preprint arXiv:1812.04606},
  year={2018}
}

@inproceedings{larochelle2008classification,
  title={Classification using discriminative restricted Boltzmann machines},
  author={Larochelle, Hugo and Bengio, Yoshua},
  booktitle={Proceedings of the 25th international conference on Machine learning},
  pages={536--543},
  year={2008}
}

@article{lee2018simple,
  title={A simple unified framework for detecting out-of-distribution samples and adversarial attacks},
  author={Lee, Kimin and Lee, Kibok and Lee, Honglak and Shin, Jinwoo},
  journal={Advances in neural information processing systems},
  volume={31},
  year={2018}
}

@article{santos2024adversarial,
  title={Adversarial training and attribution methods enable evaluation of robustness and interpretability of deep learning models for image classification},
  author={Santos, Flavio AO and Zanchettin, Cleber and Lei, Weihua and Nunes Amaral, Lu{\'\i}s A},
  journal={Physical Review E},
  volume={110},
  number={5},
  pages={054310},
  year={2024},
  publisher={APS}
}

@misc{flanagan_notmnist_mnist,
  author       = {David Flanagan},
  title        = {notMNIST-to-MNIST},
  howpublished = {\url{https://github.com/davidflanagan/notMNIST-to-MNIST}},
  note         = {GitHub repository, accessed July 9, 2026}
}

@article{falini2022review,
  title={A review on the selection criteria for the truncated SVD in Data Science applications},
  author={Falini, Antonella},
  journal={Journal of Computational Mathematics and Data Science},
  volume={5},
  pages={100064},
  year={2022},
  publisher={Elsevier}
}

@article{shannon1948mathematical,
  title={A mathematical theory of communication},
  author={Shannon, Claude Elwood},
  journal={The Bell system technical journal},
  volume={27},
  number={3},
  pages={379--423},
  year={1948},
  publisher={Nokia Bell Labs}
}

@article{plerou2002random,
  title={Random matrix approach to cross correlations in financial data},
  author={Plerou, Vasiliki and Gopikrishnan, Parameswaran and Rosenow, Bernd and Amaral, Luis A Nunes and Guhr, Thomas and Stanley, H Eugene},
  journal={Physical Review E},
  volume={65},
  number={6},
  pages={066126},
  year={2002},
  publisher={APS}
}

@article{lecun2006tutorial,
  title={A tutorial on energy-based learning},
  author={LeCun, Yann and Chopra, Sumit and Hadsell, Raia and Ranzato, M and Huang, Fujie and others},
  journal={Predicting structured data},
  volume={1},
  number={0},
  year={2006}
}

@inproceedings{
hendrycks2017a,
title={A Baseline for Detecting Misclassified and Out-of-Distribution Examples in Neural Networks},
author={Dan Hendrycks and Kevin Gimpel},
booktitle={International Conference on Learning Representations},
year={2017}
}

\end{document}